\documentclass{article}
\usepackage{spconf,amsmath,graphicx}

\usepackage{enumitem}
\usepackage{algorithmic}
\usepackage{graphicx}
\usepackage{textcomp}
\usepackage{xcolor}
\usepackage{multirow}
\usepackage{array}
\usepackage{url}
\usepackage{hyperref}
\usepackage{balance}
\usepackage{multirow}
\setlist{nosep, leftmargin=14pt}
\usepackage{tabularx}
\usepackage{mwe} 

\title{Optimizing Skin Lesion Classification via Multimodal Data and Auxiliary Task Integration}
\name{Mahapara Khurshid,
Mayank Vatsa, and Richa Singh}
\address{Indian Institute of Technology Jodhpur, India}
\begin{document}
%
\maketitle
\begin{abstract}
The rising global prevalence of skin conditions, some of which can escalate to life-threatening stages if not timely diagnosed and treated, presents a significant healthcare challenge. This issue is particularly acute in remote areas where limited access to healthcare often results in delayed treatment, allowing skin diseases to advance to more critical stages. One of the primary challenges in diagnosing skin diseases is their low inter-class variations, as many exhibit similar visual characteristics, making accurate classification challenging. This research introduces a novel multimodal method for classifying skin lesions, integrating smartphone-captured images with essential clinical and demographic information. This approach mimics the diagnostic process employed by medical professionals. A distinctive aspect of this method is the integration of an auxiliary task focused on super-resolution image prediction. This component plays a crucial role in refining visual details and enhancing feature extraction, leading to improved differentiation between classes and, consequently, elevating the overall effectiveness of the model. The experimental evaluations have been conducted using the PAD-UFES20 dataset, applying various deep-learning architectures. The results of these experiments not only demonstrate the effectiveness of the proposed method but also its potential applicability under-resourced healthcare environments.

\end{abstract}
\section{Introduction}
\label{sec:intro}
\begin{figure}[t]
 \centering
\includegraphics[scale=0.65]{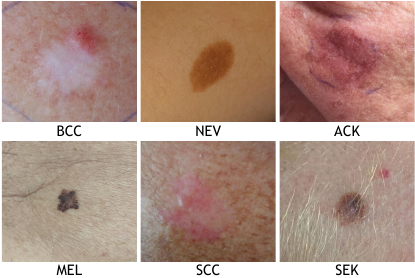}
\caption{Examples of skin lesion images  \cite{pacheco2020pad}.}
\label{fig:VA}
\vspace{-0.5cm}
\end{figure}

Skin cancer, the most prevalent form of cancer globally, has shown that early detection can significantly reduce mortality rates. Dermatologists typically use dermoscopy imaging and the ABCD rule—assessing Asymmetry, Boundary, Color, and Diameter—to diagnose skin lesions \cite{camacho2022multi}. However, this traditional method encounters challenges such as varying imaging qualities and differences in medical expertise, highlighting the necessity for computer-aided diagnostic tools.
Research in automated skin lesion classification using dermoscopic images has progressed \cite{ verdelho2022impact, khurshid2023multi}. However, as shown in Figure \ref{fig:VA}, these systems often struggle with high similarity across different classes and low similarity within the same class  \cite{wang2023intra}. To enhance accuracy, recent studies are incorporating metadata like clinical and demographic information alongside images \cite{dong2023learning, iscen2023improving}. This approach mimics the diagnostic process of dermatologists, considering more than just visual attributes, which improves the performance of automated systems. Additionally, integrating auxiliary task learning, where an additional task is learned alongside the main objective, has shown to be beneficial for generalizability \cite{he2019dme, chen2022auxiliary}.

The challenge of low inter-class similarity remains, even when using multimodal approaches that combine images with metadata. The proposed method addresses this by incorporating an auxiliary task into the classification pipeline. This involves using image features to predict super-resolution images, refining the visual details of the original images. This innovation has shown notable improvements in skin lesion classification, surpassing existing methods. This approach is particularly valuable in rural and underserved areas, where delays in accessing medical expertise and technology can lead to advanced disease progression. By utilizing smartphone-captured images and metadata, our method can be used as an effective early screening tool for these regions, enhancing the prospects of timely and accurate diagnosis.

\begin{figure*}[htbp]
\includegraphics[width=\textwidth, height=6.5cm]{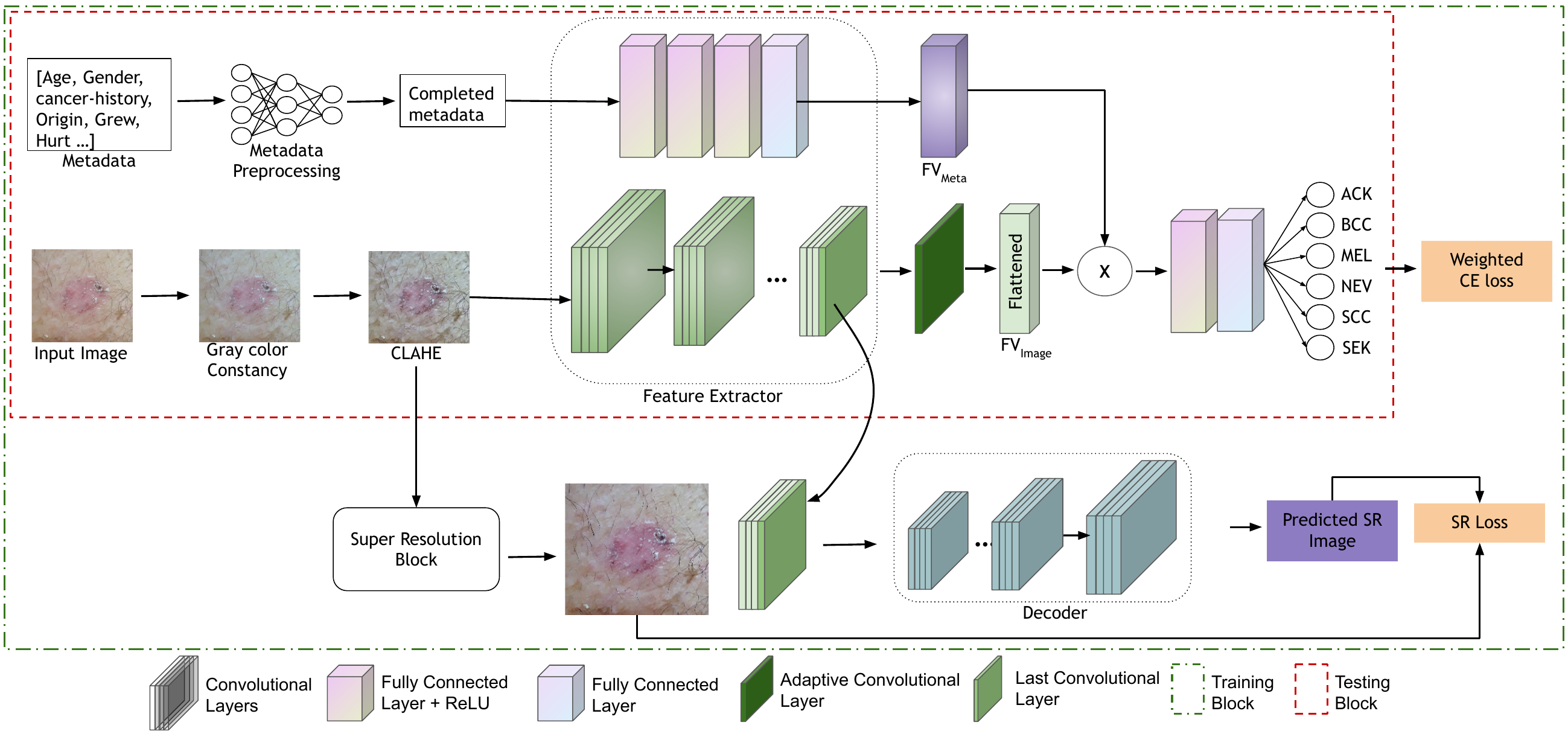}
\caption{A schematic diagram illustrating the training and testing process of the proposed approach. The auxiliary task learning guides the visual feature extractor to refine the extracted features.}
\label{fig:Block}
\vspace{-0.3cm}
\end{figure*}
\section{Proposed Approach}
This section describes the methodology of the proposed approach. The central concept underlying this research work is auxiliary task supervision, which enhances the extraction of visual features. The refined visual features are combined with metadata features to classify skin lesions. Figure \ref{fig:Block} illustrates the steps involved in the proposed method.
\newline
\textbf{Problem Formulation}:
Given $X_{img}$, a set of skin lesion images with corresponding metadata $X_{meta}$ containing various demographic and clinical features, along with diagnostic labels, y $\in$ [1...$N_{class}$], where N represents the total number of classes. The objective is to train a model that predicts y for test data ($X_{img-test}$, $X_{meta-test}$) during the inference phase.

The images are captured using smartphones under varying lighting conditions, thus incorporating some colour variations as shown in Figure \ref{fig:VA}. To handle such variations, a gray color constancy algorithm \cite{finlayson2004shades} is used to ensure colour uniformity, followed by contrast-limited adaptive histogram equalization (CLAHE) \cite{pizer1987adaptive} to enhance the contrast of the images \cite{jaisakthi2018automated}. 
The metadata includes features but they have missing values. A neural network based imputation \cite{pundhir2022towards} is applied to complete the missing information. 
\subsection{Feature Extraction}
The proposed multimodal approach consist of visual and textual feature extractors explained as follows.
\newline
\textbf{Visual Feature Extractor}: Images contain maximum information and are pivotal in enhancing the overall classification performance. To extract visual information, various Convolutional Neural Networks (CNN) are used, each varying in breadth and depth. The lower layers of the model learn features such as edges and texture, and the focus is shifted towards the shapes and patterns of the lesions. Let $\phi_{visual}$ denote the visual feature extractor, wherein the final feature maps of the selected model are flattened to create a feature vector for the input images. 
\textbf{Textual Feature Extractor}: 
To encode the metadata information, a four-layer fully connected network is employed. Each of these fully connected layers is followed by a ReLU activation function, except for the final layer. The network processes the diverse metadata features such as age, gender, bleed and cancer history, to learn the interaction within them. 
The visual and textual feature vector for image x$_{img}$ and x$_{meta}$ can be represented as equation \ref{eq:meta}.
\begin{equation}
     FV_{Image} = \phi_{visual}(x_{img}), FV_{Meta} = \phi_{meta}(x_{meta})
    \label{eq:meta}
\end{equation}
\subsection{Auxiliary Task Learning}
Various skin diseases share similar visual characteristics, leading to confusion in their classification. Hence, more distinctive features are required to enhance the classification process. These distinctive features enable the model to distinguish subtle variations that might be overlooked, thereby improving the accuracy of the classification.

The proposed approach implemented an auxiliary task to improve and refine the visual features. The process involves creating a super-resolution (SR) image from a given image by employing either an image processing technique such as bilinear or bicubic interpolation or a pre-trained deep learning model. Integrating the task helps to focus on finer image details for predicting the SR image. Since there is a shared encoder between the auxiliary task and the visual feature extractor, the visual features also get refined for the classification task. The generated SR image is a reference for comparison with the predicted SR image. 
The last feature maps generated by $\phi_{visual}$ are passed to the decoder (6-layered CNN) to predict the SR image, and the model is jointly optimized for auxiliary and classification tasks.
\subsection{Feature-level Fusion and Classification}
To integrate features from both modalities, a feature-level fusion technique based on multiplication is employed, where the textual features weigh the visual features. An adaptive convolution layer before the flattening layer is incorporated within $\phi_{visual}$ to ensure the element-wise multiplication $\odot$ of $ FV_{Image}$ and  $FV_{Meta}$, resulting in final feature vector,  $FV_{Final}$, for the classification.
\begin{equation}
     FV_{Final} =  FV_{Image}  \odot FV_{Meta}
     \label{eq:fusion}
\end{equation}
The FV$_{Final}$ is fed to the classifier consisting of two fully connected layers to give the final predicted class.

\noindent \textbf{Loss Function}
The training loss function consists of two components - weighted cross entropy loss and SR loss. The weighted cross-entropy loss, as shown in equation \ref{eq:wce}, between true label ,y, and predicted label, $\hat{y}$.
\begin{equation}
L_{wce} = - \sum_{i=1}^{N}( w_i \cdot y_i \cdot \log(\hat{y}_i) + (1 - y_i) \cdot \log(1 - \hat{y}_i) )
    \label{eq:wce}
\end{equation}
The weights '$w_i$' are computed based on the number of samples in each class.
The SR loss in equation \ref{eq:sr} gives the loss between the generated SR image and the predicted SR image.
\vspace{-0.2cm}
\begin{equation}
      L_{SR} = \frac{1}{n} \sum_{i=1}^{n} (SR_i - \hat{SR}_i)^2
    \label{eq:sr}
\end{equation}
The final loss is the weighted combination of $L_{wce}$ and $L_{SR}$.
\begin{equation}
    L_{final} = \alpha L_{wce} + \beta L_{SR}
\end{equation}
Here, $\alpha = 0.5$ and $\beta =1.0$ are adjustable factors used to bring the overall loss function in a similar range.

\section{Experimental Setup}
\textbf{Dataset Details}: This research aims to propose an approach for classifying smartphone-based skin lesions. This work utilized a publicly available dataset, PAD-UFES20 \cite{pacheco2020pad}, which comprises 2298 smartphone-captured images classified into six classes: actinic keratosis (ACK) with 730 images, basal cell carcinoma (BCC) with 845 images, melanoma (MEL) with 52 images, nevus (NEV) with 244 images, squamous cell carcinoma (SCC) with 192 images, and seborrheic keratosis (SEK) with 235 images, and their corresponding metadata. The metadata has 26 features, encompassing 21 patient clinical features (such as age, gender, anatomical location, and cancer history), a diagnostic label, and four identifiers, such as image ID and patient ID. This work followed the train-val-test split of 80:10:10 as given in \cite{pundhir2022visually}.

\noindent\textbf{Implementation details}:
For visual feature extraction, various deep neural network architectures, including VGG-13 \cite{simonyan2014very}, ResNet50 \cite{he2016deep}, MobileNet-V2 \cite{sandler2018mobilenetv2}, EfficientNet-B4 \cite{tan2019efficientnet}, and DenseNet-121 \cite{huang2017densely} are used, varying in depth and width and pretrained on the ImageNet \cite{deng2009imagenet}. The models were fine-tuned on the PAD-UFES20 dataset. For textual feature extraction, a 4-layered, fully connected network having dimensions as [64, 128, 256, 512] is used. The auxiliary task in the proposed approach is evaluated using image processing and deep learning techniques. The models were trained for 70 epochs with a batch size of 32 and utilized the SGD optimizer with an initial learning rate of 0.01. We incorporated a stepLR scheduler with a step size of 15, reducing the learning rate by 0.1 at each step. All images were resized to 224x224x3. The super-resolution block upscale the images by a factor of 2, resulting in dimensions of 448x448x3.
For classification, a weighted cross-entropy loss is used to address class imbalance and Mean Squared Error (MSE) loss is used for the auxiliary task. Various data augmentation techniques including horizontal and vertical flips, image scaling, brightness adjustments, contrast, saturation, and the addition of random noise are applied during training, to introduce variability in the data.
All the experiments were implemented using the PyTorch framework on an Nvidia DGX station.

\section{Results and Analysis}
\begin{figure}[!tbp]
\centering
\includegraphics[width=0.9\columnwidth, height =2.9cm]{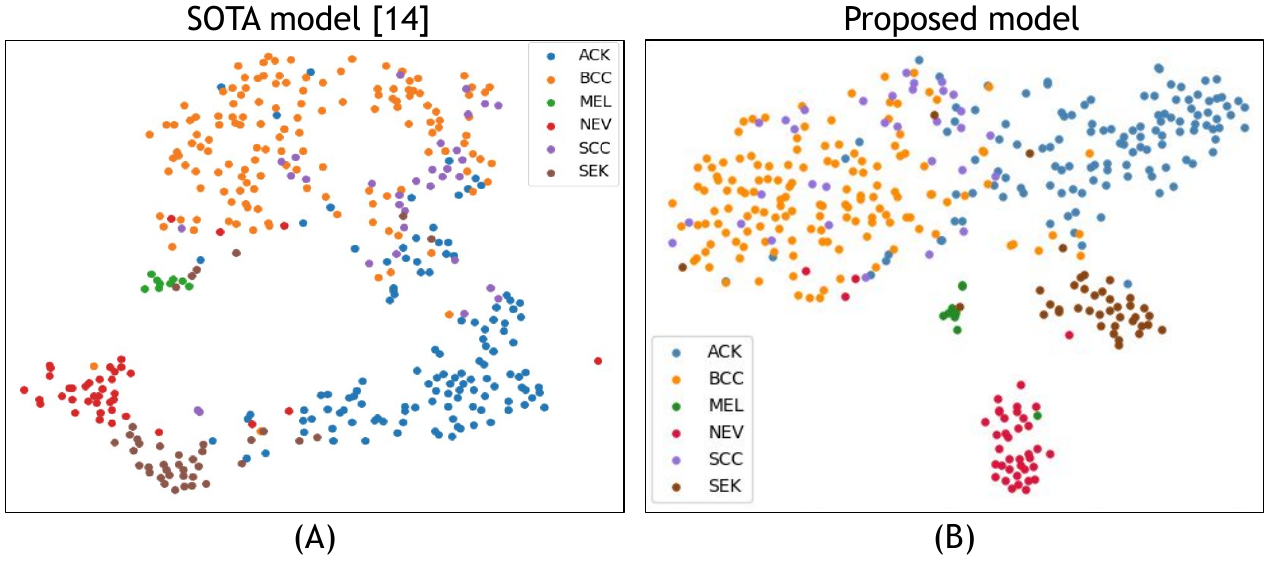}
\vspace{-0.3cm}
\caption{tSNE plots of the proposed and SOTA model.}
\label{fig:tsne}
\vspace{-0.5cm}
\end{figure}

This section presents the quantitative and qualitative results of the proposed method. The performance of the proposed approach is evaluated using balanced accuracy (BACC), classification accuracy (ACC), and area under the curve (AUC) across five distinct CNN architectures. 
Table \ref{tab:results} gives a comparative analysis of the approach against other existing methods. The proposed approach achieved a BACC of 0.832, greater than the state-of-the-art (SOTA) result of 0.806 on the PAD-UFES20 dataset. The proposed approach resulted in a BACC of 0.811 for the best-performing model in SOTA results, ResNet50, greater than the 0.806 reported in \cite{pundhir2022visually}. Table \ref{tab:results} also presents the performance of MetaNet \cite{li2020fusing}, where authors have used element-wise multiplication without an auxiliary task. It is evident that the proposed approach also outperforms their performance. This is because more discriminating features are extracted, which results in improved performance. To support this, a visual representation of the embedding space is presented, wherein samples from each class are visibly closer than the SOTA method, as demonstrated in Figure \ref{fig:tsne}. The presence of well-distinguished clusters in the visualization substantiates the performance improvement achieved by the proposed approach. The consistent improvement of performance metrics across all models demonstrates its superiority across all existing approaches. Additional evaluation metrics critical for assessing the medical diagnosis model, such as precision, recall, sensitivity, specificity, and classwise accuracy (CA) for the best-performing model, VGG13, are also computed as given in Table \ref{tab:best_model}. The specificity is greater than 90\% for all classes, indicating the accuracy of the model to identify truly negative classes. Also, the sensitivity values for each class except the SCC exceed 80\%, emphasizing the model's ability to identify true positives, which is critical for assuring reliable diagnoses. 

\begin{table}[t]
\caption{Performance comparison with existing approaches.}
\centering
\resizebox{\columnwidth}{!}{
\begin{tabular}{|l|l|ccc|}
\hline
\multicolumn{2}{|c|}{\textbf{Model}} & \textbf{ACC} & \textbf{BACC} & \textbf{AUC} \\ \hline
\multirow{5}{*}{No Metadata \cite{pacheco2021attention}} & VGG13 & 0.709 & 0.654 & 0.901 \\ \cline{2-5} 
 & ResNet50 & 0.616 & 0.651 & 0.901 \\ \cline{2-5} 
  & EfficientNet-B4 & 0.656 & 0.640 & 0.911 \\ \cline{2-5} 
 & MobileNet-V2 & 0.655 & 0.637 & 0.898 \\ \cline{2-5} 

 & DenseNet-121 & 0.636 & 0.640 & 0.893 \\ \hline
\multirow{5}{*}{Concat \cite{pacheco2020impact}} & VGG13 & 0.712 & 0.72 & 0.929 \\ \cline{2-5} 
 & ResNet50 & 0.741 & 0.728 & 0.929 \\ \cline{2-5} 
  & EfficientNet-B4 & 0.765 & 0.758 & 0.945 \\ \cline{2-5}
 & MobileNet-V2 & 0.738 & 0.741 & 0.927 \\ \cline{2-5} 
 
 & DenseNet-121 & 0.742 & 0.747 & 0.932 \\ \hline
\multirow{5}{*}{MetaNet \cite{li2020fusing}} & VGG13 & 0.749 & 0.754 & 0.937 \\ \cline{2-5} 
 & ResNet50 & 0.732 & 0.742 & 0.936 \\ \cline{2-5} 
  & EfficientNet-B4 & 0.744 & 0.737 & 0.931 \\ \cline{2-5} 
 & MobileNet-V2 & 0.700 & 0.717 & 0.922 \\ \cline{2-5} 

 & DenseNet-121 & 0.745 & 0.745 & 0.932 \\ \hline
\multirow{5}{*}{MetaBlock \cite{pacheco2021attention}} & VGG13 & 0.728 & 0.736 & 0.933 \\ \cline{2-5} 
 & ResNet50 & 0.735 & 0.765 & 0.935 \\ \cline{2-5} 
 & EfficientNet-B4 & 0.748 & 0.77 & 0.944 \\ \cline{2-5} 
 & MobileNet-V2 & 0.724 & 0.754 & 0.938 \\ \cline{2-5} 
 
 & DenseNet-121 & 0.723 & 0.746 & 0.931 \\ \hline
\multirow{5}{*}{Visual  \cite{pundhir2022visually}} & VGG13 & 0.807 & 0.77 & 0.952 \\ \cline{2-5} 
 & ResNet50 & 0.812 & 0.806 & 0.953 \\ \cline{2-5}
  & EfficientNet-B4 & 0.772 & 0.784 & 0.953 \\ \cline{2-5} 
 & MobileNet-V2 & 0.806 & 0.789 & 0.954 \\ \cline{2-5} 

 & DenseNet-121 & 0.799 & 0.779 & 0.950 \\ \hline
\multirow{5}{*}{Proposed} & VGG13 & \textbf{0.849} & \textbf{0.832} & \textbf{0.960} \\ \cline{2-5} 
 & ResNet50 & \textbf{0.848} & \textbf{0.811} & \textbf{0.967}  \\ \cline{2-5} 
  & EfficientNet-B4 & \textbf{0.833} & \textbf{0.797} & \textbf{0.967} \\ \cline{2-5}
 & MobileNet-V2 & \textbf{0.836} & \textbf{0.811} & \textbf{0.953} \\ \cline{2-5} 
 
 & DenseNet-121 & \textbf{0.822} & \textbf{0.794} & \textbf{0.962} \\ \hline
\end{tabular}}
\label{tab:results}
\vspace{-0.5cm}
\end{table}
\begin{table}[!t]
\caption{Evaluation metrics of the proposed approach.}
\centering
\resizebox{\columnwidth}{!}{
\begin{tabular}{|l|cccccc|}
\hline
\textbf{Metrics} & \textbf{ACK} & \textbf{BCC} & \textbf{MEL} & \textbf{NEV} & \textbf{SCC} & \textbf{SEK} \\ \hline
Recall & 0.88 & 0.85 & 0.89 & 0.95 & 0.50 & 0.92 \\ \hline
Precision & 0.90 & 0.88 & 0.89 & 0.93 & 0.40 & 0.95 \\ \hline
F1 Score & 0.89 & 0.87 & 0.89 & 0.94 & 0.44 & 0.94 \\ \hline
Sensitivity & 0.88 & 0.85 & 0.89 & 0.95 & 0.50  &  0.92\\ \hline
Specificity & 0.95 & 0.93 & 0.99 & 0.99 & 0.93 &  0.99\\ \hline
\begin{tabular}[l]{@{}l@{}}CA\end{tabular} & 87.70 & 85.11 & 88.88 & 95.00 & 50.00 & 92.31 \\ \hline
\end{tabular}}
\label{tab:best_model}
\vspace{-0.35cm}
\end{table}

\begin{figure}[!t]
\centering
\includegraphics[width=0.8\columnwidth, height = 4cm]{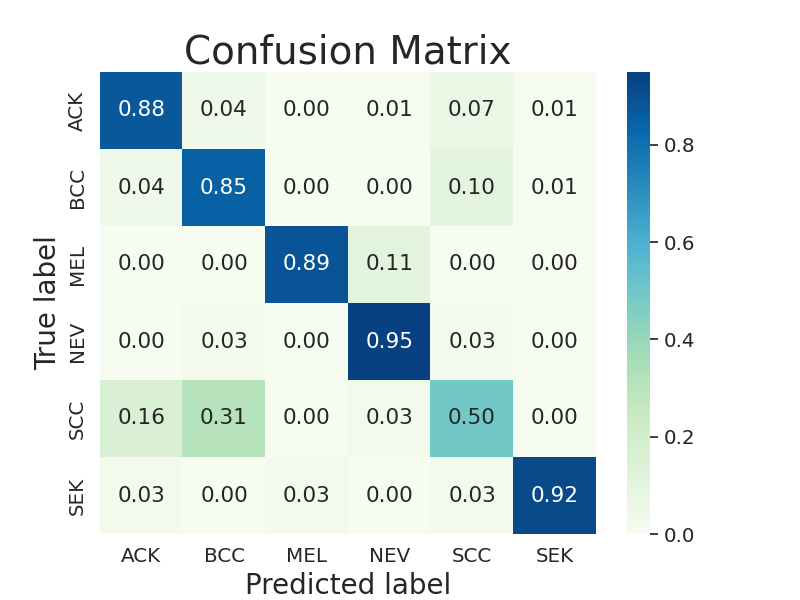}
\vspace{-0.46cm}
\caption{Confusion matrix for the best-performing model}
\label{fig:cm}
\vspace{-0.5cm}
\end{figure}

\begin{figure}[!t]
\centering
\includegraphics[width=0.8\columnwidth, height = 4cm]{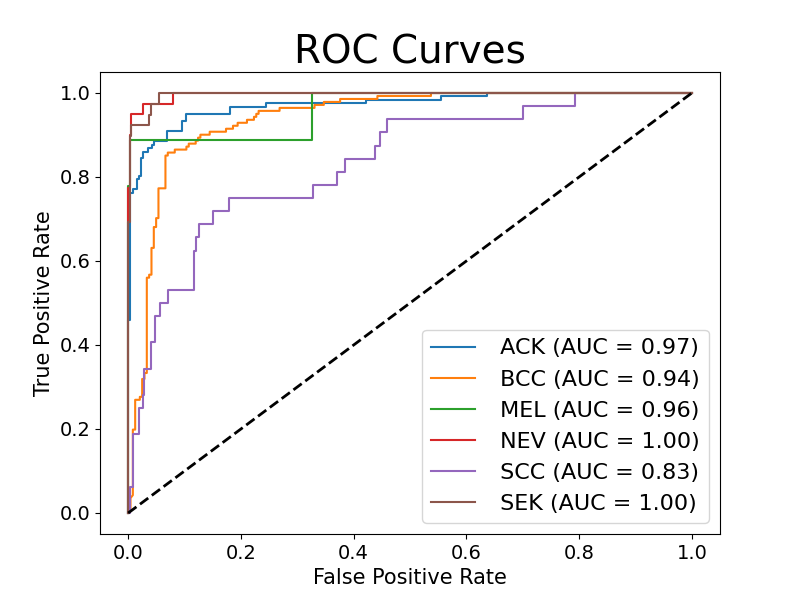}
\vspace{-0.46cm}
\caption{AUC for the best-performing model}
\label{fig:roc}
\vspace{-0.2cm}
\end{figure}

To showcase the misclassifications between different classes, the confusion matrix is shown in Figure \ref{fig:cm}, supported by the Area Under Curve (AUC) as presented in Figure \ref{fig:roc}. It is evident that most of the confusion can be seen in BCC and SCC classes. This is because these classes have minute visual differences; both are pigmented and also share a similar set of clinical features. However, it is important to note that this misclassification does not impact patient care because both classes require a further biopsy for a proper diagnosis. For further evaluation of the role of auxiliary task, we have replaced the bilinear SR technique with other techniques as reported in Table \ref{tab:srt}. The reported results also show consistent improved results for skin lesion classification.

Since the proposed approach is mainly feature-based, there is scope for improvement in extracting more distinctive features that could further solve the problem of differentiating deadly skin diseases such as MEL from minor skin diseases like ACK. Based on the reported results, the proposed method improves the skin lesion classification task and outperforms existing SOTA results for all the models.

\begin{table}[!t]
 \caption{Proposed model with varying SR techniques}
    \centering
    \begin{tabular}{|l|ccc|c|}
    \hline
         \textbf{Model} & \textbf{ACC} & \textbf{BACC} & \textbf{AUC} \\ \hline
        Bicubic & 0.849  & 0.828 & 0.965 \\ \hline
         NinaSR \cite{ninasr} & 0.838 & 0.815 & 0.963\\ \hline
         EDSR \cite{lim2017enhanced} & 0.843 & 0.821 & 0.959\\ \hline
    \end{tabular}
    \label{tab:srt}
    \vspace{-0.5cm}
\end{table}

\section{Conclusion}
This research presents a novel multimodal method for skin lesion classification that significantly enhances diagnostic accuracy by incorporating an auxiliary super-resolution image prediction task. Leveraging smartphone-captured images along with diverse clinical and demographic data, this approach significantly refines visual features, leading to enhanced performance. A thorough comparative analysis with existing methodologies reveals consistent enhancements across several key performance metrics. The distinctiveness of different lesion classes is further illustrated through a tSNE plot, providing a clear visualization of the embedding space. In future, we will extend this auxiliary task-learning technique to address different medical conditions, incorporating various modalities such as X-rays, MRI, CT scans and other related auxiliary tasks to further enhance the feature representation.

\section{COMPLIANCE WITH ETHICAL STANDARDS}
This research adheres to ethical standards, and all experiments conducted using the the publicly available dataset found at \href{https://data.mendeley.com/datasets/zr7vgbcyr2/1} {PAD-UFES20} strictly comply with the terms of use set forth by the dataset creators. The dataset utilized in this study is openly available for research purposes, and we affirm that our research does not involve any attempts at deanonymization of patients’ data. Our study prioritizes the privacy and ethical considerations associated with medical data, and we have taken measures to ensure the responsible and ethical use of the dataset in accordance with established guidelines.
\bibliographystyle{IEEEbib}
\bibliography{refs}

\end{document}